\documentclass[10pt,twocolumn,letterpaper]{article}
\usepackage{wacv}
\usepackage{times}
\usepackage{epsfig}
\usepackage{graphicx}
\usepackage{amsmath}
\usepackage{amssymb}
\usepackage{multirow}
\usepackage{makecell}
\usepackage{booktabs}
\usepackage{caption}
\usepackage{subcaption}
\usepackage{cite}

\def\wacvPaperID{156} %

\wacvfinalcopy %
\ifwacvfinal
\def\assignedStartPage{1} %
\fi
\ifwacvfinal
\usepackage[breaklinks=true,bookmarks=false]{hyperref}
\else
\usepackage[pagebackref=true,breaklinks=true,colorlinks,bookmarks=false]{hyperref}
\fi

\ifwacvfinal
\else
\fi

\begin{document}

\title{Whose hand is this? Person Identification from Egocentric Hand Gestures}

\author{Satoshi Tsutsui\\
Indiana University\\
Bloomington, IN, USA\\
{\tt\small stsutsui@indiana.edu}
\and
Yanwei Fu\\
Fudan University\\
Shanghai, China\\
{\tt\small yanweifu@fudan.edu.cn}
\and
David Crandall\\
Indiana University\\
Bloomington, IN, USA\\
{\tt\small djcran@indiana.edu}
}

\maketitle
\begin{abstract}
Recognizing people by faces and other biometrics has been extensively studied
in computer vision. But these techniques do not work for identifying
the wearer of an egocentric (first-person) camera because that person
rarely (if ever) appears in their own first-person view.
But while one's own face is not frequently visible, their hands are: in fact,
hands are among the most common objects in one's own field of view.
It is thus natural to ask whether the appearance and motion patterns
of people's hands are distinctive enough to recognize them. In this
paper, we systematically study the possibility of
Egocentric Hand Identification (EHI) with unconstrained egocentric hand
gestures. We explore several different visual cues,
including color, shape, skin texture, and depth maps to identify
users' hands.
Extensive ablation
experiments are conducted to analyze the properties of hands 
that are most
distinctive. Finally, we show that EHI can improve generalization of other tasks, such as gesture
recognition, by training adversarially to encourage these models to ignore
differences between users.
\end{abstract}

\begin{figure}[h!]
     \centering
     \begin{subfigure}[b]{0.55\linewidth}
         \centering
         \includegraphics[width=\linewidth]{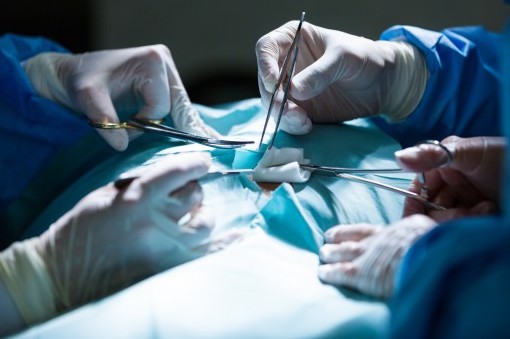}
          \caption{Medical surgery with hands \cite{photo} \label{fig:surgery} }
     \end{subfigure}
      \begin{subfigure}[b]{0.9\linewidth}
         \centering
         \includegraphics[width=\linewidth]{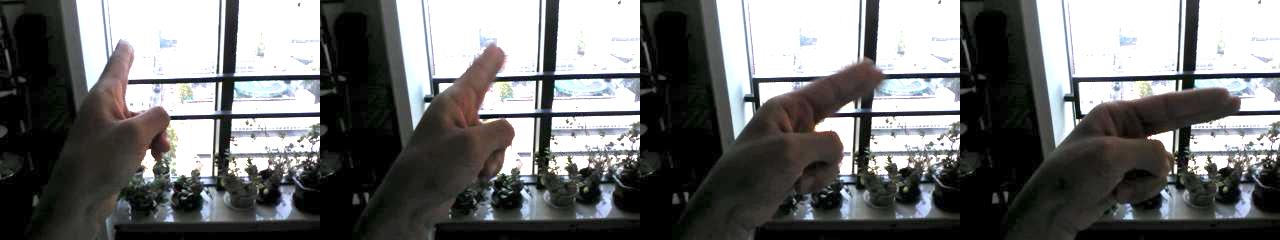}
         \caption{RGB frames}
     \end{subfigure}
         \begin{subfigure}[b]{0.9\linewidth}
         \centering
         \includegraphics[width=\linewidth]{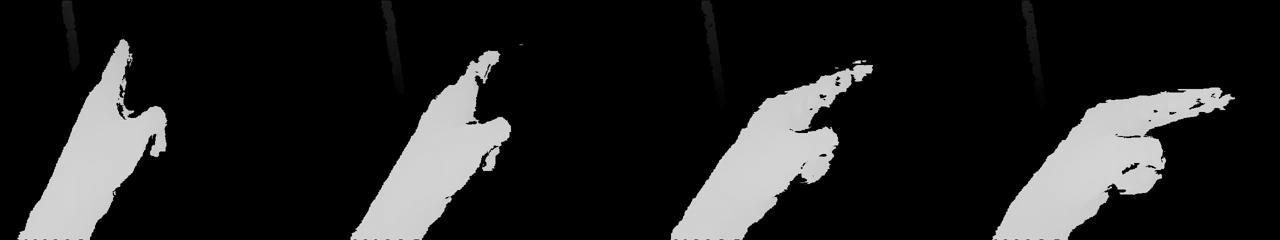}
         \caption{Depth frames}
     \end{subfigure}
         \caption{(a) If multiple people are collaborating on a task assisted by VR devices, can we tell
           whose hand is whose in the first person view?  (b,c) Sample input frames. Our results suggest that even egocentric depth data  can give evidence about who is whom.  \label{fig:teather} 
    }
\end{figure}

\section{Introduction}
Vision-based person identification has many practical applications in
safety and security applications, so developing techniques to identify
individual people is among the most-studied computer vision
problems~\cite{jain2007handbook}.  After decades of research, some
vision-based biometric technologies are now commonplace, with retina
and and palm recognition routinely used in high-security applications
such as border control~\cite{daugman2004iris}, while even consumer
smartphones feature fingerprint~\cite{kumar2013towards} and face
recognition~\cite{deng2019arcface}.  Other techniques that identify
more subtle distinguishing features, such as gait~\cite{wang2019ev} or
keystroke dynamics~\cite{banerjee2012biometric}, have also been used
in some applications.

User identification for egocentric cameras presents an interesting
challenge, however, because the face and body of the camera wearer are
rarely seen within their own first-person field of view. There is a
notable exception, however: we use our hands as a primary means of
manipulating, interacting with, and sensing the physical world around
us, and thus our hands are frequently in our field of
view~\cite{Bambach_2015_ICCV}. In fact, our own hands may be the
objects that we see most frequently throughout the entire span of our
lives and thus may even play a special role in our visual systems,
including in how children develop object
perception~\cite{smith2018developing}.

We hypothesize that it is possible to recognize individuals based only
on their hand appearances and motion patterns.  If correct, this
hypothesis would have numerous potential applications.  A VR/AR device
shared among members of a household could tailor its behavior to a
specific user's preferences. Some wearable cameras need user
authentication (e.g., verifying the wearer of a police body
camera~\cite{stross2013wearing}), and this could provide one signal.
When multiple people are interacting or collaborating on a manual task,
such as surgery (Figure \ref{fig:surgery}),
the field of view may be full of hands, and understanding the activities
in the scene would involve identifying whose hand is whose.

Some existing work has studied using properties of hands, such as
shape, to uniquely identify people, including classic papers that use
manually-engineered
features~\cite{sanchez2000biometric,yoruk2006shape}. However, these
techniques require hand images with a clearly visible palm, and are
not applicable for unconstrained egocentric videos.  

\begin{figure}[t!]
	\centering
	\renewcommand*\thesubfigure{\arabic{subfigure}}  
	\begin{subfigure}[b]{0.9\linewidth}
		\centering
		\includegraphics[width=\linewidth]{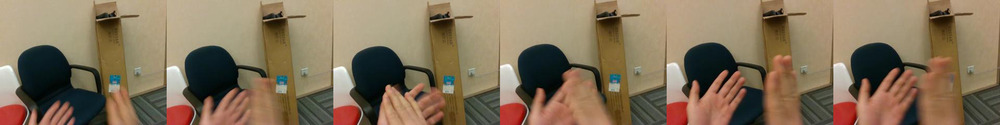}
		\caption{Original RGB.}
	\end{subfigure}
	\begin{subfigure}[b]{0.9\linewidth}
		\centering
		\includegraphics[width=\linewidth]{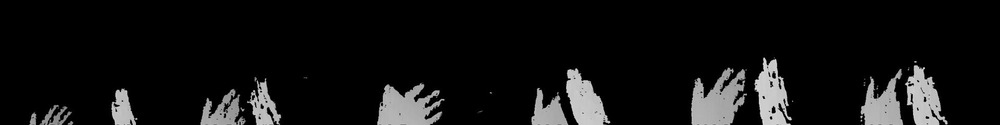}
		\caption{ Original Depth.}
	\end{subfigure}
	\begin{subfigure}[b]{0.9\linewidth}
		\centering
		\includegraphics[width=\linewidth]{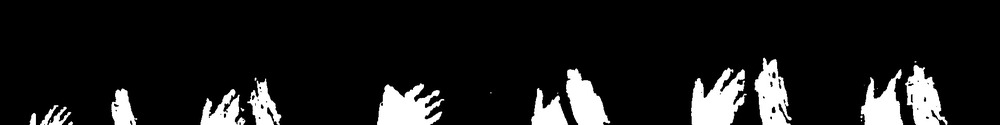}
		\caption{ Binary Hand: Binarized depth map that approximates the shapes of hands.}
	\end{subfigure}
	\begin{subfigure}[b]{0.9\linewidth}
		\centering
		\includegraphics[width=\linewidth]{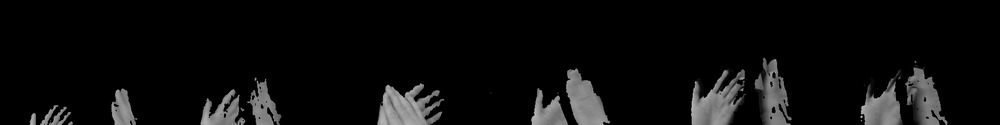}
		\caption{ Gray Hand: Grayscale images masked with hands.}
	\end{subfigure}

	\begin{subfigure}[b]{0.9\linewidth}
		\centering
		\includegraphics[width=\linewidth]{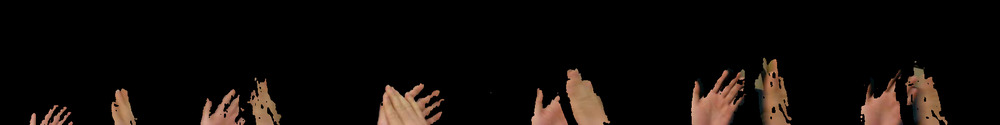}
		\caption{ Color Hand: Color images with hands.}
	\end{subfigure}
	\caption{Controlled Inputs \label{fig:input-samples}}
\end{figure}

This paper, to our knowledge for the first time, studies the task of
Egocentric Hand Identification (EHI). Since this is the first work, we
focus on establishing baselines 
and analyzing which factors of the egocentric video
contribute to the recognition accuracy. Specifically, we demonstrate
that standard computer vision techniques can learn to recognize people
based on egocentric hand gestures with reasonable accuracy even from
depth images alone.  We then conduct ablation studies to determine
which feature(s) of inputs are most useful for identification, such as
2D/3D shape, skin texture (\eg moles), and skin color.  We do this by
measuring recognition accuracy as a function of manipulations of the
input video; \eg we prepare gray-scale video of hand gestures
(Figure~\ref{fig:input-samples} (4)), and binary-pixel video composed
of hand silhouettes only (Figure~\ref{fig:input-samples} (3)), and
regard the accuracy increase between gray-scale videos and hand
silhouettes as the contribution of skin texture. Our results indicate
that all these elements (2D hand shape, 3D hand shape, skin texture,
and skin color) carry some degree of distinctive information.

We further conjectured that this task of hand identification could improve
other related tasks such as gesture recognition. However, we found
that the straightforward way of doing this -- training a multi-task
CNN to jointly recognize identity and gesture -- was actually
\textit{harmful} for gesture recognition.  While perhaps surprising at
first, this result is intuitive: we want the gesture classifier to
generalize to unseen subjects, so gesture representations should be
predictive of gestures but invariant to the person performing gestures.
We use this insight to propose an adversarial learning framework
that directly enforces this constraint: we train a primary classifier
to perform gesture recognition while also training a secondary
classifier to \textit{fail} to recognize who performs gestures. Our experiments
indicate that the learned representation is invariant to person
identity, and thus generalizes well for unseen subjects.

In summary, the contributions of this paper are as follows:

\begin{enumerate}
\item  To our
knowledge, we are the first to investigate if it is possible to identify
people based only on  egocentric videos of their hands.

\item We perform 
ablation experiments and investigate which properties of gesture videos are key
to identifying individuals.  

\item We propose an adversarial
learning framework that can improve hand gesture recognition by explicitly encouraging invariance across person identities.
\end{enumerate}

\section{Related Work}

Egocentric vision (or first-person vision) develops computer vision
techniques for wearable camera devices. In contrast to the
third-person perspective, first-person cameras have unique challenges
and opportunities for research. Previous work on egocentric computer
vision has concerned object
recognition~\cite{bertasius2017unsupervised,liu2017jointly,lee2020hand},
activity
recognition~\cite{nakamura2017jointly,zaki2017modeling,Shen_2018_ECCV,possas2018egocentric,moltisanti2017trespassing,Damen_2018_ECCV,patra2019ego,arora2018stabilizing},
gaze
prediction~\cite{Huang_2018_ECCV,Li_2018_ECCV,zhang2017deep,tavakoli2019digging},
video
summarization~\cite{penna2017summarization,HO_2018_ECCV,silva2018weighted},
head motion signatures~\cite{poleg2014head}, and human pose
estimation~\cite{yuan20183d,jiang2017seeing}.  In particular, many
papers have considered problems related to hands in egocentric video,
including hand pose estimation~\cite{mueller2017real,garcia2018first},
hand segmentation~\cite{urooj2018analysis,Bambach_2015_ICCV}, handheld
controller tracking~\cite{Pandey_2018_ECCV}, and hand gesture
recognition~\cite{cao2017egocentric}. However, to our knowledge, no
prior work has considered egocentric hand-based person
identification. The closest work we know of is extracting user
identity information from optical
flow~\cite{hoshen2016egocentric,issharing20}, which is
complementary to our work --- integrating optical flow into our
study would be future work.

In a broad sense, using hands to identify people has been
well-studied, particularly for fingerprint
identification~\cite{kumar2013towards}, but it is not realistic to
assume that we can extract fingerprints from egocentric video.
Outside of biometrics, some classic papers have shown the potential to
identify subjects based on hand shapes and geometries. Sanchez-Reillo
et al. use palm and lateral views of the hand, and define features
based on widths and heights of predefined key
points~\cite{sanchez2000biometric}.  Boreki et al. propose 25
geometric features of the finger and palm~\cite{boreki2005hand}, while
Yoruk et al. use independent component analysis on binary hand
silhouettes~\cite{yoruk2006shape}. However, these studies use very
clean scans of hands with the palm clearly visible, often with special
scanning equipment specifically designed for the
task~\cite{sanchez2000biometric}.  Views of hands from wearable cameras are
unconstrained, much noisier, and capture much greater diversity of hand poses,
so these classic methods are not applicable.  Instead, we build on modern
computer vision to learn features effective for this task in a
data-driven manner.

We are aware of two studies that apply modern computer vision for hand
identification. Mahmoud~\cite{afifi201911kHands} applies CNNs for
learning representations from controlled hand images with the palm
clearly visible.  Uemori et al. use multispectral images of skin patch
from hands, and train 3D CNNs to classify the
subjects~\cite{uemori2019skin}. The underlying assumption for this
approach is that each individual has distinctive skin spectra due to
the unique chromophore concentrations within their
skin~\cite{fitzpatrick1988validity}. Although this method does not
have any constraints for hand pose, it requires a multispectral
sensors, which are not usually available
in consumer devices.  Moreover, it assumes images of the hands are
clear, not blurry, and with good lighting conditions, which are not
practical assumptions for egocentric vision.

\section{Methodology}

Our goal is to investigate if it is possible to identify a camera
wearer from an egocentric video clip of them performing hand
gestures. Our goal is not to engineer a new method but to investigate
whether a simple end-to-end technique can learn to perform this
task. Specifically, we build upon a standard convolutional neural
network~\cite{hara2018can} for video classification, and train an
end-to-end subject classifier from video clips.  Our approach uses the
raw inputs from RGB or depth sensors directly, and intentionally does not
employ hand-pose or hand-mesh recognition because we do not want our
method to depend on the recognition results of other tasks (which
would complicate the method and make it vulnerable to failures of
those subtasks).

\subsection{Egocentric Hand Identification (EHI)}\label{sec:hand-id}

Our focus is to understand 
the influence of various sources of evidence --- RGB, depth, etc. ---
on the classification decisions. This is important due to the
data-driven nature of deep learning and difficulty in collecting
large-scale egocentric video datasets that are truly representative
of all people and environments: it is possible for a classifier
to \textit{cheat} by learning bias of the training data (e.g., if all
gestures of a subject are recorded in the same room, the classifier
could memorize the background). We try to avoid this by making sure
the training and testing videos are recorded in different places, but
we also try to identify the types of features (\eg hand shapes, skin
texture, skin color) that are important for CNNs to identify the
subjects.  In order to factor out each element, we ablate the input
video and gradually increase information starting from the silhouettes
of hands to the full-color images of hands. In the remainder of this
section, we discuss the above-mentioned points in detail, starting
from the RGB and Depth inputs available to us.

\newcommand{\xhdr}[1]{\vspace{4pt}\noindent {\textbf{#1.}}}

\xhdr{RGB} We have RGB video clips of gestures against various backgrounds.
The RGB data implicitly captures 2D hand
shape, skin texture, and skin color. We show
a sample clip in Figure \ref{fig:input-samples} (1), containing not
only hands but also a doll. In fact, the same doll appears in other
gesture videos of the same subject, which is problematic if the person classification model
learns to rely on this background information. This is not just a
dataset problem, but a practical problem for possible real-world applications
of egocentric vision: a user may record videos in their room to register
their hands in the system but still want the device to recognize
hands elsewhere.  To simulate this point, we train on indoor videos
and evaluate on outdoor videos.  

\xhdr{Depth} We also have depth video synchronized with the RGB clips
(Figure \ref{fig:input-samples} (2)).  The depth images contain
information about 3D hand shape in addition to shapes of objects in
the background.  A clear advantage of using depth is that it allows
for accurate depth-based segmentation of the hands from the background,
just by thresholding on distance from the camera.  Although less
sensitive to the background problem, there is still a chance that
depth sensors capture the geometries of the background (\eg rooms). In
order to eliminate background, we apply Otsu's binarization
algorithm~\cite{otsu1979threshold} to separate the background and
foreground. It is reasonable to assume that hands are distinctively
closer to the device, so the foreground mask corresponds to binary hand
silhouettes -- an approximation of the hand shapes without 3D
information.  These hand silhouettes are a starting point for our
ablative study.

\xhdr{Binary Hand} We obtain  binary hand silhouettes by
binarizing the depth maps as discussed above. We show an example in
\ref{fig:input-samples} (3).  This only contains  hand shape
information and is the input with the least information in our study.
We prepare several other inputs by adding more information to this
binary video and use the accuracy gain to measure the contribution of additional
information, as described below.

\xhdr{3D Hand} We apply the binary mask to the depth images and
extract the depth corresponding to the hand region only. The accuracy increase
from Binary Hand is a measure of the importance of 3D hand shape
for identifying the subjects.

\xhdr{Gray Hand} We apply the binary mask to the grayscale frame
converted from RGB; see example in Figure
\ref{fig:input-samples} (4). This adds hand texture information to
the 2D Hands. The accuracy gain from Binary Hand indicates the
importance of textures of hands (including moles and nails).

\xhdr{Color Hand} We extract the hand from RGB frames, eliminating
the background. We show an example in Figure
\ref{fig:input-samples} (5). The accuracy gain from Gray Hand is the
contribution of skin color.

\xhdr{3D Color Hand} This is the combination of all available
information where hands are extracted from RGBD frames. The accuracy
indicates synergies of 2D hand shape, 3D hand shape, skin texture, and
skin color.

\subsection{Improving gesture recognition}\label{sec:advtrain}

We hypothesize that our technique is not only useful for identifying
people from hands, but also for assisting in other gesture recognition
problems. This hypothesis is based on results on other problems that
show that multi-task training can perform
each individual task better, perhaps because the multiple tasks help
to learn a better feature representation.  A naive way to implement
the idea of jointly estimating gesture and identity is to train
gesture classification together with subject classifier, sharing the
feature extraction CNN. Then, we can minimize a joint loss
function,
\begin{equation}
\min_{ (F,C_g ,C_p )} \left( \mathcal{L}_{g} + \mathcal{L}_{p} \right),
\label{eq:multi}
\end{equation}
where $F,C_g,C_p, \mathcal{L}_{g}, \mathcal{L}_{p}$ are the shared CNNs
for feature extraction, the classifier to recognize the gestures, the
classifier to recognize identity, the loss for gesture
recognition, and the loss for recognizing subjects, respectively.

However, and somewhat surprisingly, we empirically found that minimizing this joint loss \textit{decreases}
the accuracy of gesture recognition. Our interpretation is that
because  the gesture classification is trained and tested on 
disjoint subjects, learning a representation that is \textit{predictive} of person identity 
is  harmful in classifying the gestures performed by
new subjects. We hypothesized that a model trained for the ``opposite'' task,
learning a representation \textit{invariant to} identity,
 should better generalize to  unseen subjects. This can be
expressed as a min-max problem of the two tasks,
\begin{equation}
\min_{ (F,C_g )} \max_{C_p}  \left( \mathcal{L}_{g} + \mathcal{L}_{p} \right). 
\label{eq:adv}
\end{equation}

Intuitively, this equation encourages the CNN to learn the joint
representation ($F$) that can predict gestures ($C_g$) but cannot
predict who performs the gestures ($C_p$). This min-max problem is the
same as that used in adversarial domain
adaptation~\cite{ganin2014unsupervised}, because the gesture classifier is
trained adversarially with the subject classifier. Following the original
method~\cite{ganin2014unsupervised}, we  re-write
equation~\ref{eq:adv} into an alternating optimization of two equations,
\begin{equation}\label{eq:adv_1}
    \min_{ (F,C_g )} \left( \mathcal{L}_{g} - \lambda \mathcal{L}_{p} \right)  
\end{equation}
\begin{equation}\label{eq:adv_2}
     \max_{C_p}  \mathcal{L}_{p},
\end{equation}
where $\lambda$ is a hyper-parameter. 

\section{Experiments}

We perform our primary experiments on
the EgoGesture~\cite{cao2017egocentric,zhang2018egogesture} dataset, which
is a large dataset of hands taken both indoors and outdoors. We
also perform secondary experiments on the relatively small 
Green Gesture (GGesture)~\cite{tejo2018ismar} dataset.

\xhdr{Implementation Details} We use ResNet18~\cite{he2016resnet} with 3D
 convolutions~\cite{hara2018can}. We
 initialize the model with  weights pretrained from the Kinetics
 dataset, and fine-tune with mini-batch stochastic gradient descent
 using Adam~\cite{adam} with batch size of 32 and  initial
 learning rate of 0.0001. We iterate over the dataset for 20 epochs
 and decrease the learning rate by a factor of 10 at the 10th
 epoch. The spatiotemporal input size of the model is $112 \times 112
 \times 16$, corresponding to width, height,
 and frame lengths respectively. In order to fit  arbitrary input
 clips into this resolution, we resize the spatial resolution into $171
 \times 128$ via bilinear interpolation, then apply random crops for training and center crops for
 testing. For the temporal dimension, we train with randomly-sampled clips of 16
 consecutive frames and test by averaging the prediction of 16
 consecutive sliding windows with 8 overlapping frames. If the
 number of frames is less than 16, we pad to 16 by repeating the
 first and last frames.

\subsection{Experiments on EgoGesture}
The EgoGesture~\cite{cao2017egocentric,zhang2018egogesture} dataset includes
24,161 egocentric video clips of 83 gesture classes performed by 50
subjects, recorded both indoors and outdoors. Our task is to classify each
video into one of  50 subjects. As discussed in
Sec~\ref{sec:hand-id}, we use indoor videos for training, and outdoor
videos for evaluation, so that we can prevent leaking user identities based 
on the backgrounds. The dataset has 16,373 indoor clips, which are used
for training, and 7,788 outdoor clips of which  we use 3,892 for
validation and 3,896 for testing.

\begin{table*}[tb!]
    \begin{center}
    \begin{tabular}{lcccccccc}
    \toprule
    \multicolumn{1}{c}{} & \multicolumn{5}{c}{Information}    & &\multicolumn{2}{c}{Accuracy} \\
    \cmidrule{2-6}
        \cmidrule{8-9}
    \multicolumn{1}{c}{}                       & 2D Shape    & 3D Shape & Skin Texture & Skin Color & Background & & EgoGesture            & GGesture           \\

    \midrule
    RGB   &\checkmark&-&\checkmark&\checkmark&\checkmark&& 6.29 & 61.09 \\
    Depth &\checkmark&\checkmark&-&-&\checkmark & & 11.47 & - \\
    \midrule
    Binary Hand &\checkmark&-&-&-&-&& 11.11 & 51.13 \\
    3D Hand &\checkmark&\checkmark&-&-&-&&12.04 & - \\
    Gray Hand &\checkmark&-&\checkmark&-&-& &14.22 & 61.09 \\
    Color Hand &\checkmark&-&\checkmark&\checkmark&-&& 18.53 & 64.71 \\
    \midrule
    3D Color Hand &\checkmark&\checkmark&\checkmark&\checkmark&-& &19.53 & - \\
    \bottomrule
    \end{tabular}
    \end{center}
    \caption{Subject recognition accuracy (\%) for each ablated input on EgoGesture dataset and GGesture dataset.\label{tbl:results-egogesture-person}}
\end{table*}

\newcommand{\halfimg}{429px}
\newcommand{\inputfigsize}{1}
\begin{figure}[h!]
	\centering
	\renewcommand*\thesubfigure{\arabic{subfigure}}  
	\begin{subfigure}[b]{0.9\linewidth}
		\centering
		\includegraphics[trim = \halfimg{} 0 \halfimg{} 0, clip, width=\inputfigsize\linewidth]{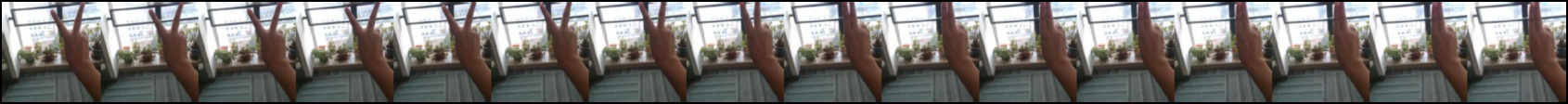}
		\caption{A training gesture clip recorded indoor.}
	\end{subfigure}
	\begin{subfigure}[b]{0.9\linewidth}
		\centering
		\includegraphics[trim = \halfimg{} 0 \halfimg{} 0, clip, width=\inputfigsize\linewidth]{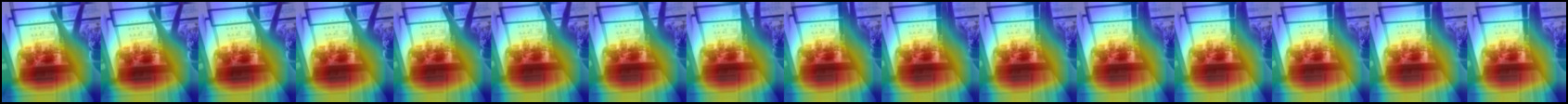}
		\caption{CAM showing that prediction is cuing on background.}
	\end{subfigure}
	\begin{subfigure}[b]{0.9\linewidth}
		\centering
		\includegraphics[trim = \halfimg{} 0 \halfimg{} 0, clip, width=\inputfigsize\linewidth]{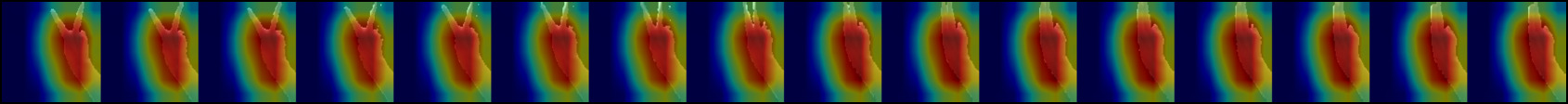}
		\caption{CAM focuses on hands if we mask out the background.}
	\end{subfigure}
	\begin{subfigure}[b]{0.9\linewidth}
		\centering
		\includegraphics[trim = \halfimg{} 0 \halfimg{} 0, clip, width=\inputfigsize\linewidth]{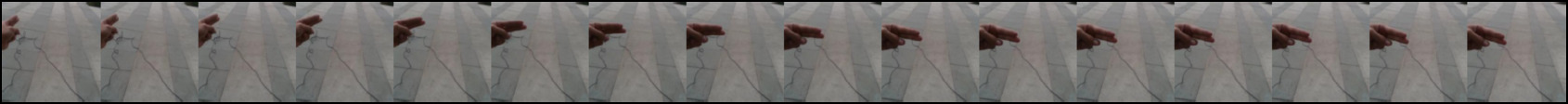}
		\caption{A test clip recorded outdoors.}
	\end{subfigure}

	\begin{subfigure}[b]{0.9\linewidth}
		\centering
		\includegraphics[trim = \halfimg{} 0 \halfimg{} 0, clip, width=\inputfigsize\linewidth]{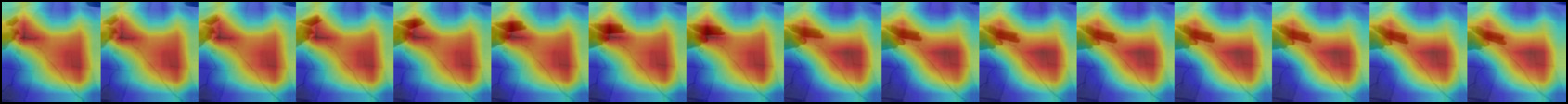}
		\caption{CAM focusing on background.}
	\end{subfigure}
	
	\begin{subfigure}[b]{0.9\linewidth}
		\centering
		\includegraphics[trim = \halfimg{} 0 \halfimg{} 0, clip, width=\inputfigsize\linewidth]{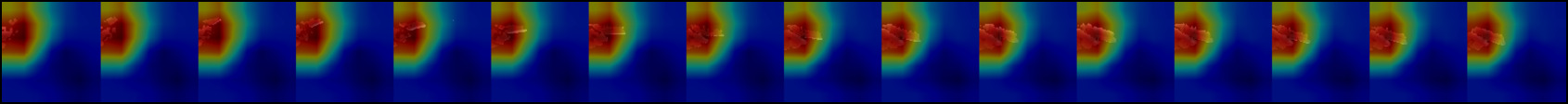}

		\caption{CAM focuses on hands when background is masked out.}
	\end{subfigure}
	
	\caption{Class Activation Maps (CAMs) \label{fig:input-cam}}
\end{figure}

\begin{figure}
     \centering
     \begin{subfigure}[b]{0.26\linewidth}
         \centering
         \includegraphics[width=\linewidth]{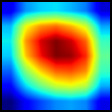}
          \caption{RGB}
     \end{subfigure}
     \hfill
     \begin{subfigure}[b]{0.26\linewidth}
         \centering
         \includegraphics[width=\linewidth]{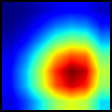}
         \caption{Color Hand}
     \end{subfigure}
     \hfill
     \begin{subfigure}[b]{0.26\linewidth}
         \centering
         \includegraphics[width=\linewidth]{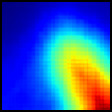}
          \caption{Actual Hand}
     \end{subfigure}
     \vspace{-6pt}
    \caption{(a) Average CAM produced by the model trained from raw images with hand and background. (b) Average CAM produced by the model trained from the images only with hand. (c) Average of the actual hand masks. Overall, CAM by the hand-only model has its peak closer to the actual hand location.  \label{fig:ave-cam}}
\end{figure}

\subsubsection{Subject Recognition Results}
We summarize the subject recognition results in Table
\ref{tbl:results-egogesture-person}. The original inputs of RGB and
Depth can achieve accuracies (\%) of 6.29 and 11.47. Because a
random guess baseline is 2\%, our results indicate the potential to
recognize people based on egocentric hand gestures. In order to
understand where this information comes from, we factor the input into different components
 and discuss the results below.

The accuracy on Binary Hand is 11.11\%. This input contains only
hand silhouettes, indicating that it is possible to recognize a person's
identity to some degree using 2D shape alone. This is the starting
point of our analysis as we gradually add additional
information. Adding depth information to the binary
mask (3D hand) increases the accuracy to 12.04\%. This indicates that
3D information, including the distance from a camera to hands,
contributes to better identify the subjects. The gray-scale images of
hands (Gray Hands) achieve an accuracy of 14.22\%, suggesting that the
texture of skin carries some information. RGB images of hands
(Color Hand) can achieve an accuracy of 18.53\%, suggesting that
skin color also conveys information about personal identity.

The results so far show that 2D shapes of hands, 3D shapes of hands,
skin texture, and skin color carry information about the identity of a
subject.  We can combine all of them with RGBD frames masked with hand
segmentation maps (3D Color Hand), and this achieves the highest
accuracy of 19.53\%. This indicates that each property contributes at least some
additional, complementary information.

\subsubsection{Background overfitting via CAM Visualization}

In terms of unmodified inputs, the accuracy  of RGB (6.29\%) is
significantly worse than depth (11.47\%). This is somewhat surprising
given that our ablative study shows that RGB images only with hands
have much higher accuracy (18.53\%) and that skin color and texture
information are  important cues. We suspect the reason lies in the way
we design the task: because we train on indoor videos and test on
outdoor videos, the CNNs can overfit on background specific to the
subjects.

To test this hypothesis, we perform experiments with class
activation maps (CAM)~\cite{zhou2016learning} to visualize the image
evidence used for making classification decisions.
CAM is  a linear combination of the feature maps
produced by the last convolutional layer, with the linear coefficients
proportional to the weight in the last classification layer. CAM
highlights regions that the network is relying on. We show a sample
clip each from training set and test set along with CAMs from RGB
and Color Hand models in Figure \ref{fig:input-cam}. As we
expected, if the backgrounds are not removed, the classifier seems
to cue on the background instead of hands. These are just two random
samples, so in order to provide more robust results, we sample 1000
clips from the test set, compute the average CAM image, and compare it
with the average image of binary hand masks. As shown in Figure
\ref{fig:ave-cam}, the hand-only input produces an average CAM image
whose peak is closer to the peak of the actual hand mask. This
observation supports our hypothesis that raw RGB frames lead the
model to overfit to the background, causing the test accuracy drop.

\begin{table}[tb!]
    \begin{center}
    \begin{tabular}{lcc}
    \toprule
    Input & Acc & Diff from 3D CNN\\
    \midrule
    Binary Hand & 12.09 & $+0.98$ \\
    3D Hand & 11.14 & $-0.90$ \\
    Gray Hand & 12.86 & $-1.36$ \\
    Color Hand & 16.84 & $-1.69$ \\
    \bottomrule
    \end{tabular}
    \end{center}
    \caption{Difference in subject recognition accuracy when swapping 3D convolution with 2D convolution. \label{tbl:motion-matters}}
\end{table}

\begin{figure}
     \centering
     \begin{subfigure}[b]{\linewidth}
         \centering
         \includegraphics[width=0.85\linewidth]{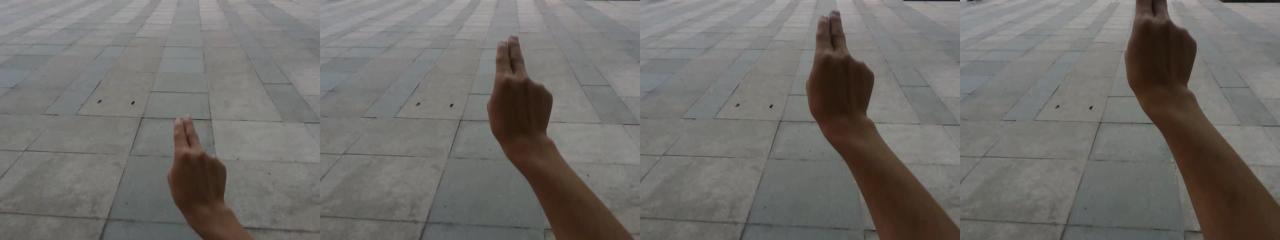}
          \caption{Gesture 80. A gesture that is easy to recognize subject.}
     \end{subfigure}
     \begin{subfigure}[b]{\linewidth}
         \centering
         \includegraphics[width=0.85\linewidth]{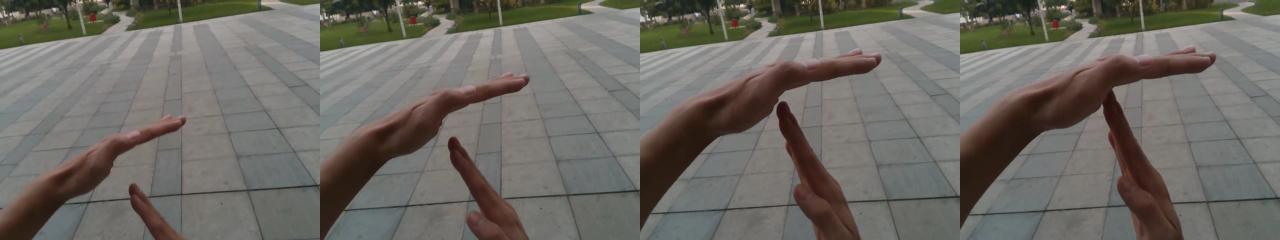}
          \caption{Gesture 36. A gesture that is hard to recognize subject.}
     \end{subfigure}
     \begin{subfigure}[b]{\linewidth}
         \centering
         \includegraphics[width=0.85\linewidth]{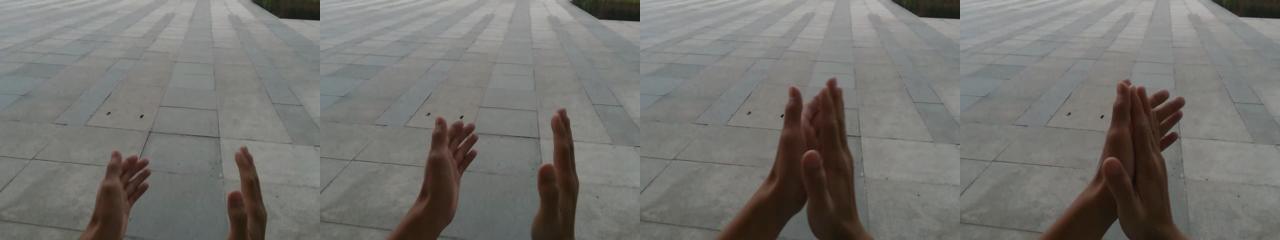}
          \caption{Gesture 52. A gesture that is hard to recognize subject.}
     \end{subfigure}
    \caption{Sample gestures that are (a) easy and (b,c) hard for recognizing subjects. \label{fig:sample-gesture-cls}}
\end{figure}

\subsubsection{Importance of Motion Information}
We use 3D convolutional networks for all experiments, so all models
internally use the information of hand motion. However, it is
interesting to measure how much motion information contributes to
accuracy. To investigate this, we swap all 3D convolutions with 2D
convolutions,  extract the 2D CNN feature for each frame, and use
the average feature vector as a representation of the clip. We compare
the accuracy of this model with the 3D CNNs. Table
\ref{tbl:motion-matters} shows the results. For 3D Hand, Gray Hand,
and Color Hand, the accuracy  drops around 1 point, indicating the
importance of motion information. However, to our surprise, the
accuracy increases 0.98\% for the input of Binary Hand. This indicates
that when we only capture 2D hand shapes, considering the temporal
feature is not helpful and possibly confusing.
We speculate that because 2D hand shapes are essentially only edges of
hands, they do  not carry enough information to distinguish complex hand motion.

\subsubsection{Effect of Gesture Class}
We suspect that some gestures are more suitable to classifying subjects than others.
Therefore, we investigate accuracy based on gesture
types. This analysis is possible because the dataset was originally
proposed for gesture recognition. We compute the subject recognition
accuracy per gesture class and show it in Figure
\ref{fig:accuracy_per_gesture} in Supplementary Material. While the highest accuracy is 29.79\%
with gesture 80, the lowest is 10.64\% with gesture 36 and 52. We show
 samples on gestures 80, 36, and 52 in Figure
\ref{fig:sample-gesture-cls}. The easy case (gesture 80) uses a
single hand with the back of the hand clearly visible, while the 
difficult ones (gesture 36 and 52) use both hands with lateral
views, making it harder to identify the subject.

\subsubsection{Effect of Clip Length}
\begin{figure}
    \begin{center}
    \includegraphics[width=0.9\linewidth]{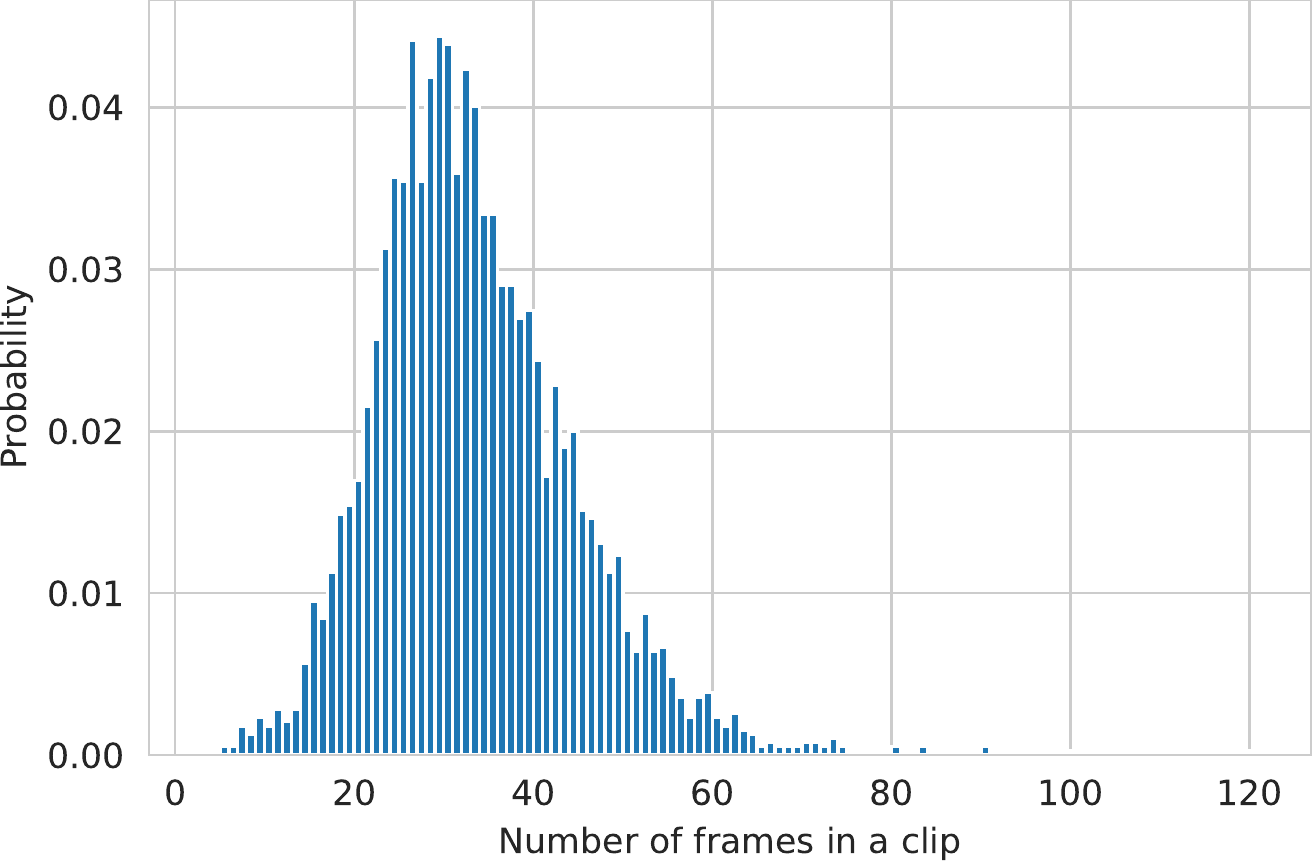}
    \end{center}
    \caption{Histogram of number of frames per video in the EgoGesture test set. \label{fig:hisogram}}
\end{figure}

Figure~\ref{fig:hisogram}
shows a histogram of clip length in the test set. The shortest clip
only has three frames while the longest  has 122 frames. The mean
length is 33.1 and the median is 32. Does the length of clip affect
the recognition accuracy? To investigate this, we divide the test set
into short, medium, and long clips based on the 25 (26 frames) and 75 percentile (39 frames)
of the length distribution.
We use 3D Color Hand as input and show the accuracy based on this split in Table
\ref{tbl:length-accuracy}.  The short, medium, and long clips have 
accuracies of 19.13\%, 20.55\%, and 17.84\%, respectively.
This result was not expected because we hypothesized  
that longer clips would be easier  because they contain more information.
We speculate
 that since the CNN is trained with a fixed-length input (16 consecutive
frames), if the clip length is too long (longer than 39 -- more than
twice that of the fixed input lengths), the CNN cannot effectively capture
key information compared to medium-length clips.

\begin{table}[tb!]
    \begin{center}
    \begin{tabular}{lccc}
    \toprule
     & Short & Medium & Long \\
    \midrule
    Accuracy   & 19.13 & 20.55 & 17.84 \\
    \bottomrule
    \end{tabular}
    \vspace{-3pt}
    \end{center}
    \caption{Accuracy (\%) over the clip length for subject recognition. We divide the test clips into short, medium, and long with the boundaries of 25\% quantile and 75\% quantile. 
    \label{tbl:length-accuracy}
    }
\end{table}

\begin{table}[tb!]
    \begin{center}
    \begin{tabular}{lcc}
    \toprule
     Input & Seen Gestures & Unseen Gestures (Diff)\\
    \midrule
    Binary Hand & 11.34 & 7.40 (-3.94) \\
    3D Hand & 11.49 & 10.24 (-1.25)\\
    Gray Hand & 14.82 & 12.01 (-2.81)\\
    Color Hand & 19.50 & 14.85 (-4.65) \\
    \bottomrule
    \end{tabular}
    \vspace{-3pt}
    \end{center} 
    \caption{Subject recognition accuracy (\%) drop for the seen and unseen gestures. We train the model with only half the available gestures and compute the test accuracy on seen and unseen gestures separately to see the generalization ability in terms of unseen hand pose. \label{tbl:useen-accuracy}}
\end{table}

\begin{figure}
	\begin{center}
		\includegraphics[width=0.9\linewidth]{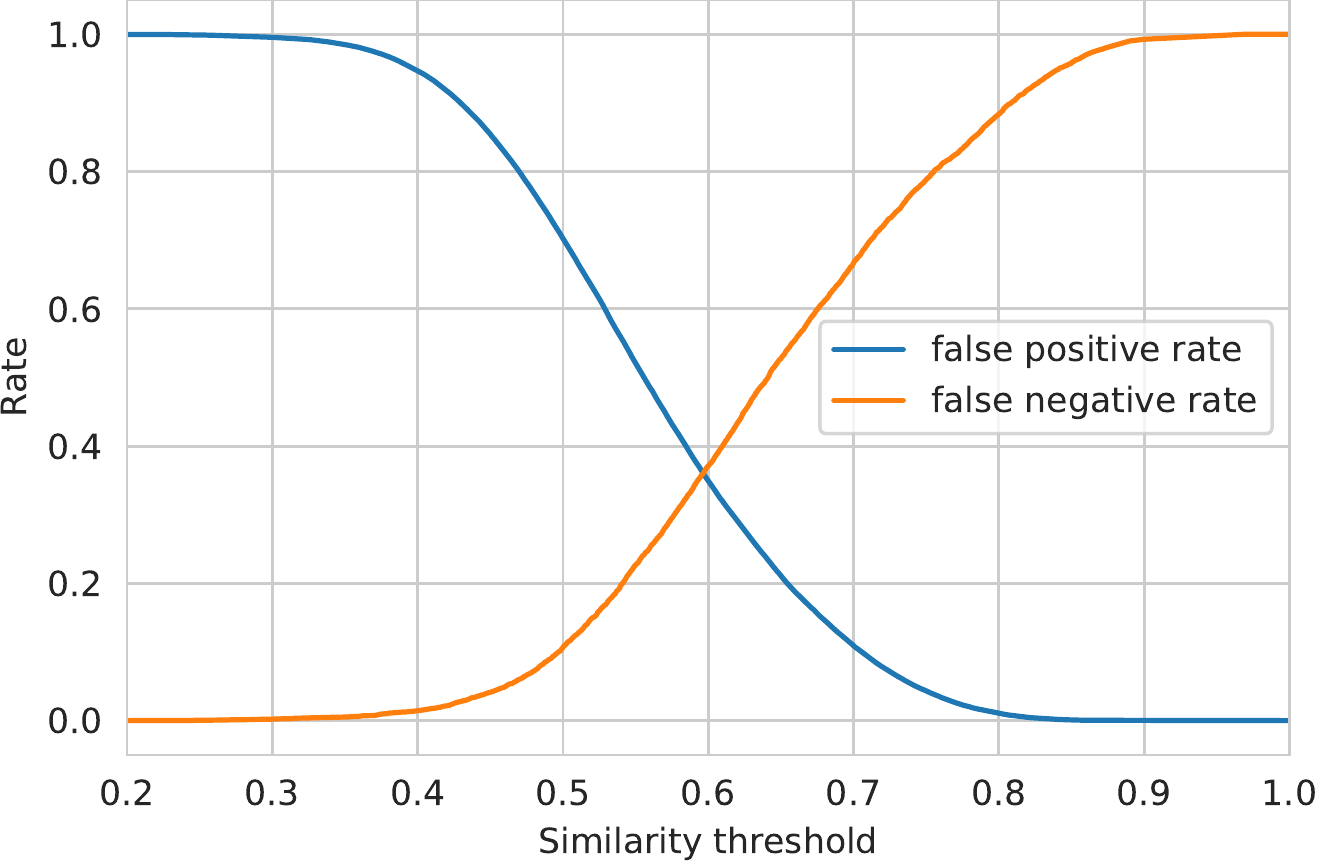}
	\end{center}
	\caption{Trade-off between false positive rate and false negative rate for gesture-based user verification settings. \label{fig:thresholds}}
\end{figure}

\subsubsection{Subject Recognition for Unseen Gestures}
Our experiments so far divide the dataset based on its recorded place
(indoor for training and outdoor for testing), and all gesture classes
appear in both training and test sets. Another question is how well the model
generalizes to unseen gestures. To answer this,
we subsampled the training set by choosing 
half the gestures (exact split provided in the Supplementary
Material), and compute test accuracy for seen and
unseen gestures separately. As shown in Table
\ref{tbl:useen-accuracy}, the accuracy drops by 3.94, 1,25, 2.81,
and 4.65 percentage points for Binary Hand, 3D Hand, Gray Hand, and Color Hand
respectively. As expected, it becomes more difficult to recognize the
subjects if the hands are in unseen poses.

\begin{figure*}[t]
	\centering
	\renewcommand*\thesubfigure{\arabic{subfigure}}  
	\begin{subfigure}[b]{\linewidth}
		\centering
		\includegraphics[width=\linewidth]{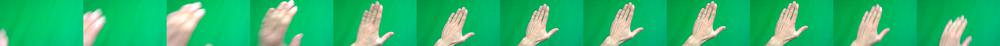}
		\caption{Sample frames}
	\end{subfigure}
	\begin{subfigure}[b]{\linewidth}
		\centering
		\includegraphics[width=\linewidth]{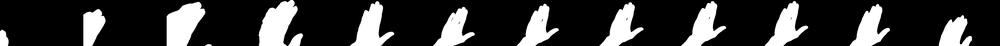}
		\caption{Extracted hand masks by removing the green chromakey background.}
	\end{subfigure}
	\vspace{-2mm}
	\caption{  Green Gesture Dataset\label{fig:greengesture} }
\end{figure*}

\subsubsection{Gesture Based Verification}
So far, the task is to recognize the subject that appeared in the
training set, but a practical scenario could be user
verification: given a pair of gesture clips, 
judge if they are from the same subject or not, even if the subject
has not been seen before.
To test this, we
use 30 subjects for training, 10 for validation, and 10 for testing,
and provide the exact split in Supplementary Material.  With this
split, we learn the representation by training on a classification task,
but for evaluation, we perform gesture-based user verification by
thresholding the cosine similarity of clip pairs. An evaluation pair
consists of an indoor clip and an outdoor clip performing the same
gesture. This amounts to 58,888 pairs of video clips with a heavy
class imbalance, where only 5,950 pairs are positive. To incorporate
this imbalance into the evaluation, we report Equal Error Rate (EER)
where the threshold is set to have the same false-positive rate and
false-negative rate.  We evaluate with the input of 3D Color Hand and
obtain an EER of 36.01\%.  We plot the trade-off between false-positive
rate and false-negative rate for different thresholds in Figure
\ref{fig:thresholds}. We also plot the ROC curve in Figure
\ref{fig:roc} in Supplementary Material.

\subsubsection{Gesture Recognition Results}
In addition to recognizing the subject, our model potentially could benefit
existing gesture recognition tasks as well. For these experiments, we use the
data split defined in the original dataset where train/val/test have
disjoint subjects because the model is expected to generalize to
unseen people. We experiment with two ways to train the gesture
recognition model with the auxiliary subject classifier as described
in Sec. \ref{sec:advtrain}, and summarize the results in Table
\ref{tbl:results-egogesture-gesture}. The first way is to jointly
train the gesture classifier and subject classifier while sharing the
internal CNN representation as a multi-task learning task. Unfortunately,
this method has an accuracy  drop from the single-task training
both for RGB (from 88.42\% to 86.91\%) and for Depth (from 89.35\% to
86.64\%). This suggests that the representations predictive of subjects
are  harmful to classify gestures, and the representations
invariant to subjects are better. This is intuitive given that the
gesture recognition model is expected to generalize to unseen
subjects. Therefore we realize this representation with
adversarial training and observe the accuracy  improves both
for RGB (from 88.42\% to 89.51\%) and for Depth (from 89.35\% to 89.66\%).

\begin{table}[tb!]
    \begin{center}
    \begin{tabular}{lcc}
    \toprule
     & RGB   & Depth \\
    \midrule
    Gesture only & 88.42 &  89.35 \\
    Joint with Subject-ID & 86.91 & 86.64\\
    Adversarial with Subject-ID &  \textbf{89.51} & \textbf{89.66}\\ 
    \bottomrule
    \end{tabular}
    \end{center}
    \caption{Gesture recognition accuracy (\%) on EgoGesture dataset\label{tbl:results-egogesture-gesture}.}\vspace{-8pt}
\end{table}  

\subsection{Subject Recognition on GGesture Dataset}
We also perform subject classification experiments on the Green Gesture
(GGesture)~\cite{tejo2018ismar} dataset. This dataset has about 700
egocentric video clips of 10 gesture classes performed by 22
subjects. It has RGB clips without depth, and recorded only indoors.
However, unlike the EgoGesture dataset, the background is always a green
screen for chroma key composition, so we can easily extract the hand
masks. Each subject has three rounds of recordings so we use them for
train/val/test split, resulting in 229/216/221 clips, respectively. We
show the results in Table~\ref{tbl:results-egogesture-gesture} next to
the EgoGesture results. Because GGesture  does not have depth
information, we only ablate the input with Binary Hand, Gray Hand, and
Color Hand, which have accuracies of 51.13\%, 61.09\%, and 64.71\%
respectively. The accuracy gain corresponds to the contribution of 2D
shape of hands, skin texture, and skin colors, respectively.

\section{Discussion and Conclusion}
We have presented a CNN-based approach to recognize subjects from
egocentric hand gestures. To our knowledge, this is the first study to
show the potential to identify subjects based on ego-centric views of
hands. We also perform ablative experiments to investigate the
properties of input videos that contribute to the recognition
performance. The experiments shows that hands shape in 2D and 3D, skin
texture, skin colors, and hand motions are all keys to identify a person's
identity. Moreover, we also show that training gesture classifiers
adversarially with subject identity recognition can improve the gesture
recognition accuracy.

Our work has several limitations.  First, while our hand masks remove
the background and most other objects, it is possible that the model
still cues on users' hand accessories (\eg rings) to identify the
user. We are aware that at least three subjects out of 50 in the
EgoGesture dataset wear rings in some videos. Nevertheless, this is
not an issue when we use depth modality or only hand shapes, and we
note that the accuracy for those cases is greater than 10\% over 50
subjects where random guess accuracy is only 2\%.  Second, the
performance is far from perfect, with a verification error rate of
around 36\%, which means more than one out of every three
verifications is wrong.  However, this work is a first step in solving
a new problem, and first approaches often have low performance and
strong assumptions; object recognition 15 years ago was evaluated with
six class classification --- which was referred to as ``six diverse
object categories presenting a challenging mixture of visual
characteristics''~\cite{fergus2003object} --- as opposed to the
thousands today. Nonetheless, we experimentally showed that our task
can benefit hand gesture recognition.
We hope our work inspires more work in hand-based identity
recognition, especially in the context of first-person vision.

\xhdr{Acknowledgments} This work was supported in part by the National
Science Foundation (CAREER IIS-1253549) and by Indiana University
through the Emerging Areas of Research Initiative \textit{Learning:
  Brains, Machines and Children}.
  
{\small
\bibliographystyle{ieee_fullname}
\bibliography{references}
}

\clearpage
\section{Supplementary Material}
\subsection{Subject Recognition Accuracy per Gesture}

\begin{table}[h]
    \centering
\scalebox{0.98}{
\begin{tabular}{rr}
\toprule
 Label Index &    Accuracy \\
 \midrule
  52 &  10.64 \\
  36 &  10.64 \\
  16 &  12.77 \\
  68 &  12.77 \\
  61 &  12.77 \\
  14 &  12.77 \\
  60 &  14.58 \\
  45 &  14.89 \\
  21 &  14.89 \\
  17 &  14.89 \\
  15 &  14.89 \\
  31 &  14.89 \\
  65 &  14.89 \\
  71 &  14.89 \\
  53 &  14.89 \\
  77 &  15.22 \\
  13 &  17.02 \\
  43 &  17.02 \\
  26 &  17.02 \\
  25 &  17.02 \\
  24 &  17.02 \\
  23 &  17.02 \\
  46 &  17.02 \\
  49 &  17.02 \\
  51 &  17.02 \\
  63 &  17.02 \\
  10 &  17.02 \\
  75 &  17.02 \\
  83 &  17.02 \\
  73 &  17.39 \\
  39 &  17.39 \\
  42 &  17.39 \\
  11 &  19.15 \\
  19 &  19.15 \\
  8 &  19.15 \\
  30 &  19.15 \\
  50 &  19.15 \\
  62 &  19.15 \\
  29 &  19.15 \\
  40 &  19.15 \\
 \midrule
\end{tabular}

\begin{tabular}{rr}
\toprule
 Label Index &    Accuracy \\
 \midrule
 56 &  19.15 \\
  9 &  19.15 \\
  69 &  19.15 \\
  3 &  19.15 \\
  6 &  19.15 \\
  12 &  19.57 \\
  78 &  19.57 \\
  41 &  20.83 \\
  81 &  20.83 \\
  54 &  21.28 \\
  55 &  21.28 \\
  59 &  21.28 \\
  58 &  21.28 \\
  7 &  21.28 \\
  74 &  21.28 \\
  22 &  21.28 \\
  27 &  21.28 \\
  2 &  21.28 \\
  38 &  21.28 \\
  79 &  21.28 \\
  67 &  23.40 \\
  4 &  23.40 \\
  76 &  23.40 \\
  5 &  23.40 \\
  72 &  23.40 \\
  33 &  23.40 \\
  64 &  23.40 \\
  82 &  23.40 \\
  47 &  23.40 \\
  48 &  23.40 \\
  44 &  23.40 \\
  57 &  23.40 \\
  18 &  23.40 \\
  1 &  23.40 \\
  70 &  23.91 \\
  66 &  23.91 \\
  20 &  25.53 \\
  32 &  25.53 \\
  35 &  25.53 \\
  34 &  25.53 \\
  28 &  27.66 \\
  37 &  27.66 \\
  80 &  29.79 \\
  \bottomrule
\end{tabular}
}
    \caption{Subject recognition accuracy as a function of gesture class, sorted by accuracy. This is the corresponding data for Figure \ref{fig:accuracy_per_gesture}}.
    \label{tab:accuracy_per_gesture}
\end{table}

\subsection{Dataset Split for Subject Recognition for Unseen Gestures}
We select the following gestures for training: 2,  4,  6,  8,  10,  12,  14,  16,  18,  20,  22,  24,  26,  28,  30,  32,  34,  36,  38,  40,  42,  44,  46,  48,  50,  52,  54,  56,  58,  60,  62,  64,  66,  68,  70,  72,  74,  76,  78,  80, and 82.

\subsection{Dataset Split for Gesture Based Verification}
We use 30 subjects for training, 10 for validation, and 10 for testing. The subjects for train is 3, 4, 5, 6, 8, 10, 15, 16, 17, 20, 21, 22, 23, 25, 26, 27, 30, 32, 36, 38, 39, 40, 42, 43, 44, 45, 46, 48, 49, and 50. The validation is 2, 9, 11, 14, 18, 19, 28, 31, 41, and 47. The testing is 1, 7, 12, 13, 24, 29, 33, 34, 35, and 37.    

\begin{figure}
	\begin{center}
		\includegraphics[width=\linewidth]{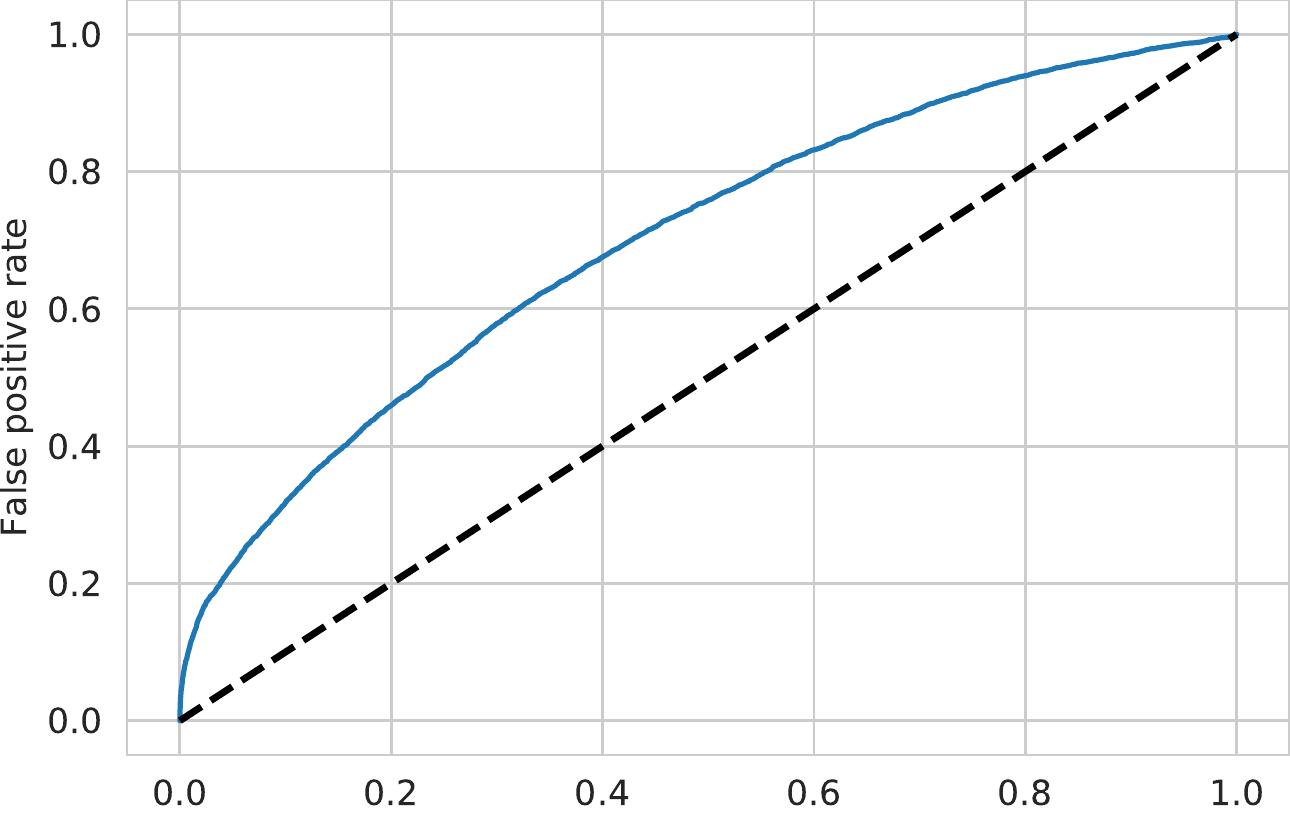}
	\end{center}
	\caption{ROC curve for gesture-based user verification\label{fig:roc}}
\end{figure}

\begin{figure*}
    \begin{center}
    \includegraphics[width=\linewidth]{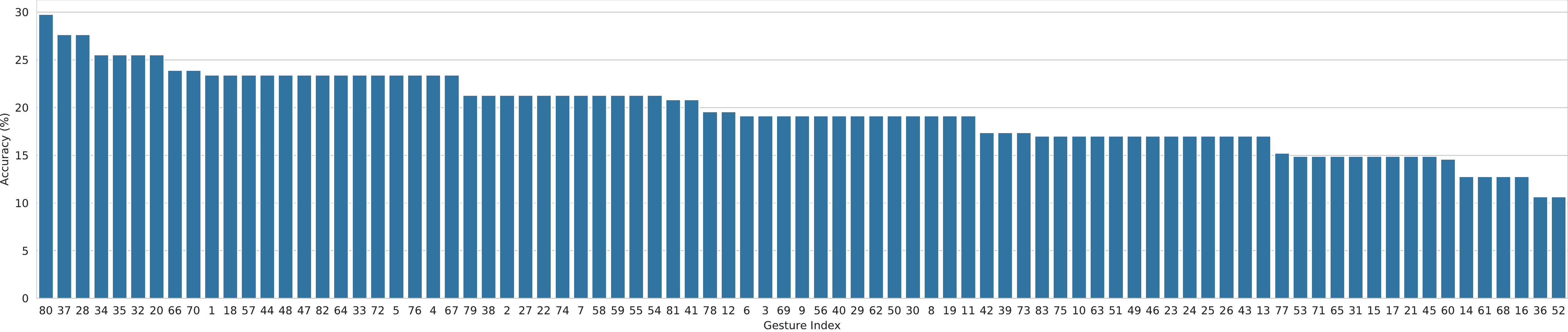}
    \end{center}
    \caption{Subject recognition accuracy as a function of gesture class, sorted by accuracy. Some gestures allow  subjects to be more easily recognized than others. Table in Supplementary Material shows the original data. \label{fig:accuracy_per_gesture}}
\end{figure*}

\end{document}